\documentclass[lettersize,journal]{IEEEtran}
\usepackage{amsmath,amsfonts}
\usepackage{algorithmic}
\usepackage{algorithm}
\usepackage{array}
\usepackage[caption=false,font=small,labelfont=sf,textfont=sf]{subfig}
\usepackage{textcomp}
\usepackage{stfloats}
\usepackage{url}
\usepackage{verbatim}
\usepackage{graphicx}
\usepackage{cite}
\usepackage{tabularx}
\usepackage{multirow}
\usepackage{booktabs}
\usepackage{color}
\newtheorem{problem}{Problem}

\begin{document}

\title{Counterfactual Data Augmentation with Denoising Diffusion for Graph Anomaly Detection}

\author{Chunjing Xiao,~Shikang Pang,~Xovee Xu,~Xuan Li,~Goce Trajcevski,~Fan Zhou
\thanks{Manuscript received April 30 2023; revised January 22 2024; accepted May 13 2024. This work was supported in part by the Natural Science Foundation of China under Grant 62176043. {(\emph{Corresponding author: Xovee Xu and Fan Zhou}.)}}


\thanks{C. Xiao and S. Pang are with the School of Computer and Information Engineering, Henan University, and also with the Henan Key Laboratory of Big Data Analysis and Processing, Henan University, Kaifeng 475004, China (email: chunjingxiao@gmail.com, pangsk0604@henu.edu.cn).}%
\thanks{X. Xu and F. Zhou are with the University of Electronic Science and Technology of China, Chengdu, Sichuan 610054, China (email: xovee@std.uestc.edu.cn, fan.zhou@uestc.edu.cn).}%
\thanks{X. Li is with the National Key Laboratory of Fundamental Science on Synthetic Vision, Sichuan University, Chengdu 610065, China (email: lixuanlmw@stu.scu.edu.cn).}%
\thanks{G. Trajcevski is with the Department of Electrical and Computer Engineering, Iowa State University, Ames, IA 50011, USA (email: gocet25@iastate.edu).}%

}

\markboth{IEEE Transactions on Computational Social Systems}%
{Xiao \MakeLowercase{\textit{et al.}}: Counterfactual Data Augmentation with Denoising Diffusion for Graph Anomaly Detection}

\IEEEpubid{0000--0000/00\$00.00~\copyright~2023 IEEE}

\maketitle

\begin{abstract}
  A critical aspect of Graph Neural Networks (GNNs) is to enhance the node representations by aggregating node neighborhood information. However, when detecting anomalies, the representations of abnormal nodes are prone to be averaged by normal neighbors, making the learned anomaly representations less distinguishable. To tackle this issue, we propose CAGAD -- an unsupervised Counterfactual data Augmentation method for Graph Anomaly Detection -- which introduces a graph pointer neural network as the heterophilic node detector to identify potential anomalies whose neighborhoods are normal-node-dominant. For each identified potential anomaly, we design a graph-specific diffusion model to translate a part of its neighbors, which are probably normal, into anomalous ones. At last, we involve these translated neighbors in GNN neighborhood aggregation to produce counterfactual representations of anomalies. Through aggregating the translated anomalous neighbors, counterfactual representations become more distinguishable and further advocate detection performance. 
  The experimental results on four datasets demonstrate that CAGAD significantly outperforms strong baselines, with an average improvement of 2.35\% on F1, 2.53\% on AUC-ROC, and 2.79\% on AUC-PR.
\end{abstract}


\section{Introduction}

\IEEEPARstart{A}{nomaly} detection on graph aims to detect nodes that present abnormal behaviors and significantly deviate from the majority
of nodes~\cite{ma2021comprehensive,liu2022benchmarking}.
Anomaly detection has numerous high-impact applications in various domains, such as abnormal user detection~\cite{shen2022trust,yang2021mining} and fraud behavior detection~\cite{cui2021remember,hu2020understanding}.
To identify anomalies in graph-structured data, it is of paramount importance to learn expressive and distinguishable node representations by modeling the information from both node attributes and graph topology.
Among many approaches that tried different learning methods for 
anomaly detection, graph neural networks (GNNs)~\cite{chen2019exploiting,xu2019powerful} have gained significant attention due to their effectiveness and flexibility~\cite{xu2019powerful,velivckovic2018graph}.

\textbf{Motivation:} One fundamental assumption in GNNs is that similar nodes (w.r.t. node attributes and their labels) have a higher tendency to connect to each other than the dissimilar ones.
Hence, GNNs typically utilize the message-passing scheme to learn node representations by aggregating node attributes iteratively from their local neighborhoods.
However, GNN-based graph anomaly detection models confront a critical over-smoothing issue~\cite{tang2022rethinking,chai2022can} inherent to 
GNNs -- namely, when aggregating information from the neighborhoods, the representations of anomalous nodes are averaged by normal nodes, as the number of normal nodes is much larger than the number of anomalies.
This, in turn, can make the anomaly representations less distinguishable. Hence, the benign neighbors of anomalies might attenuate the suspiciousness of anomalies, resulting in a poor detection performance.


\IEEEpubidadjcol

Several GNN models are proposed to remedy this issue, 
mainly falling in two categories: 
(1) \textit{Re-sampling} methods balance the number of samples by over-sampling the minority class, or under-sampling the majority class~\cite{dou2020enhancing,liu2021pick,liu2020alleviating}. (2) \textit{Re-weighting} methods assign different weights to different classes or even different samples by cost-sensitive adjustments or meta-learning-based methods~\cite{wang2019semi,cui2020deterrent,liu2021intention}.
Despite considerable performance improvements, they still suffer from certain shortcomings:
($i$) Most of the models rely on massive labeled samples to obtain a good performance. 
However, labeled anomalies are scarce and obtaining them is costly and time-consuming, as anomalies are typically rare data instances in most practical applications~\cite{pang2021deep,ma2021comprehensive}.
($ii$) These methods mainly manipulate the training data for building a robust classification model. However, for the anomaly detection task, the phenomenon of imbalanced neighbors also exists in practical (testing) data. Ignoring the manipulation of test data might result in significant performance degradation.

\textbf{Our contributions:} To overcome these drawbacks, we propose to manipulate neighbors of anomalous nodes,
so that the learned anomaly representations are distinguishable from the normal ones.
Performing direct manipulation is very difficult since we do not have label information about which nodes are anomalies. Fortunately, 
the anomaly detection task only has two classes of nodes, 
thus it is feasible to identify whether a target node has heterophily-dominant neighbors -- i.e., most of its neighbors have different class labels with the target node~\cite{he2022block,yang2022graph}. 
We refer to these nodes as \textit{heterophilic} nodes, a common occurrence in anomaly detection tasks due to class imbalance. Our intention is to manipulate the neighbors of heterophilic nodes to enhance node representations.

\begin{figure}[t]
    \centering
    \includegraphics[width=\linewidth]{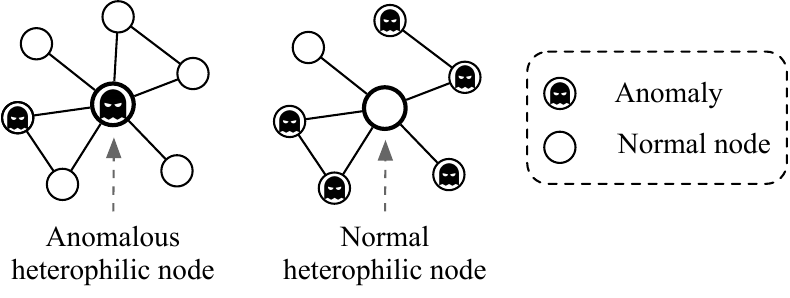}

    \caption{Nodes with heterophily-dominant neighbors.}
    \label{fig:heterophilic-node}
\end{figure}

Heterophilic nodes can be either normal or anomalous, as illustrated in \figurename~\ref{fig:heterophilic-node}. 
For an anomalous heterophilic node, most of its neighbors are normal nodes, and for a normal heterophilic node, most of its neighbors are anomalies \cite{he2022block}.
Since the labels of heterophilic nodes are unknown, we treat anomalous and normal heterophilic nodes equally, i.e., for each heterophilic node, we translate a part of its neighbors into anomalous ones, and involve these translated nodes for neighborhood aggregation.
Specifically, for anomalous heterophilic nodes, translating a part of their neighbors (which are probably normal) into anomalies would enhance the neighborhood aggregation for learning distinguishable node representations. Likewise, for normal heterophilic nodes, translating their neighborhoods (which are probably anomalous) into anomalies would have only a small impact on the neighborhood aggregation, since we do not change the labels of these neighborhoods. The selection of translated nodes can be guided by a ranking strategy according to their similarities to the heterophilic node.

Based on this idea, we present CAGAD -- Counterfactual data Augmentation for Graph Anomaly Detection --
which follows a \textit{detection $\rightarrow$ translation $\rightarrow$ representation} scheme: (\textit{i}) we first introduce a graph pointer neural network as the heterophilic node detector to identify nodes with heterophily-dominant neighbors.
(\textit{ii}) Then, for each heterophilic node,
we translate a part of its neighbors into anomalous ones using a probabilistic anomaly generator, which is powered by a denoising diffusion probabilistic model (DDPM) \cite{ho2020denoising}.
The generator can extract distinct features of anomalies, and these features are regarded as the conditions to be imposed on the generative process to create new anomalous neighbors for the heterophilic node.
(\textit{iii}) Finally, 
the generated anomalous neighbors are involved in a counterfactual GNN that is built
by incorporating the anomaly generator into a graph attention network (GAT). Counterfactual GNN aggregates the information of new unseen neighbors and produces counterfactual node representations.

The detection, translation, and aggregation processes in CAGAD are all performed in an unsupervised way without labeled data.
Unlike many of the traditional anomaly detection methods, our proposed CAGAD can be applied to test data for enhancing the node representations and will not impact the nodes' identification information (i.e., the generated counterfactual representations still have the same labels as the original ones, as changing a few neighbors or edges will not impact the node's identification (label) information ~\cite{zhu2021graph,zheng2021generative}).
Through this scheme, our model learns more distinguishable  representations for anomalous nodes and alleviates the over-smoothing issue of GNNs in graph anomaly detection.
The main contributions of our work are as follows:

\begin{itemize}
    \item We propose a novel Counterfactual data Augmentation method for Graph Anomaly Detection (CAGAD), which can produce counterfactual representations by aggregating unseen neighbors to enhance the representations of anomalies in an unsupervised learning way.

    \item We design a probabilistic anomaly generator powered by denoising diffusion models. It iteratively extracts distinct features of anomalies, which are involved in the generative process to transfer neighbors into anomalous ones.

    \item We propose a counterfactual GNN to generate counterfactual node representations by aggregating the translated neighbors rather than the original ones, which can alleviate the smooth aggregation problem of GNNs.

    \item Extensive experiments on four public anomaly detection datasets showed that CAGAD significantly improves the detection performance compared to state-of-the-art baselines\footnote{Source code is released at \url{https://github.com/ChunjingXiao/CAGAD}}.
\end{itemize}

The rest of this paper is organized as follows. 
Section \ref{preliminary}  describes the problem formulation and necessary backgrounds.
Section \ref{sec:method} presents the details of the our CAGAD model. 
Section \ref{sec:experiment} compares our model with baseline approaches on four datasets. 
Section \ref{sec:relate} reviews the literature related to anomaly detection, graph data augmentation and diffusion models. 
Finally, we conclude this work with future directions in Section \ref{sec:conclusion}.



\section{Preliminaries}
\label{preliminary}

We now formulate the graph anomaly detection problem and give the necessary background information on the denoising diffusion probabilistic models, which are used to generate new anomalous neighbors.

\subsection{Problem Formulation}
\label{problem}

\begin{table}[!ht]
    \centering
    \caption{List of Notations.}
    \label{tab:Notations}
    \begin{tabular}{l|l}
    \hline
        \toprule
        Notation & Description \\
        \midrule
        $v$ & The node vector of a graph \\
        $u$ & The attention score of a node \\
        $l$ & A hidden layer in the graph attention networks \\
        $\eta/\alpha$ & A given threshold\\
        $\mathbf{h}/\textbf{z}/\mathbf{d}/\mathbf{e}$ & The feature vector of a node \\        
        \midrule
        $\mathcal{V}$ & The set of nodes of a graph \\
        $\mathcal{E}$ & The set of edges of a graph \\
        $G$ & A graph $G = (\mathcal{V},\mathcal{E},\textbf{X})$ \\
        $\textbf{X}$ & The node feature matrix that $G$ may have  \\
        $\textbf{A}$ & The adjacent matrix of a graph \\
        $\mathbb{R}$ & The set of real numbers \\
        $\mathbf{H}/\textbf{Z}$ &  The feature matrix of a graph \\

        \bottomrule
    \end{tabular}
\end{table}


Following the commonly used notations, we use calligraphic fonts, bold lowercase letters, and bold uppercase letters to denote sets (e.g., $\mathcal{V}$), vectors (e.g., $\textbf{x}$), and matrices (e.g., $\textbf{X}$), respectively. In general, an attributed network can be represented as $G = (\mathcal{V},\mathcal{E},\textbf{X})$, where $\mathcal{V}=\{v_1,v_2,...,v_n\}$ denotes the set of nodes,  $\mathcal{E}=\{e_1,e_2,...,e_m\}$  denotes the set of edges, and $\textbf{X}=\{\textbf{x}_1,\textbf{x}_2,...,\textbf{x}_n\} \in \mathbb{R}^{n \times h}$ denotes the $h$-dimensional attributes of $n$ nodes. A binary adjacency matrix $\textbf{A} \in \mathbb{R}^{n \times n}$ is the structural information of the attributed network, where $\textbf{A}_{i,j}=1$ if there is a link between nodes $v_i$ and $v_j$; $\textbf{A}_{i,j}=0$ otherwise. Accordingly, the problem of anomaly detection on an attributed graph is defined as follows:

\begin{problem}[Anomaly Detection]
Let $\mathcal{V}_a$ and $\mathcal{V}_n$ be two disjoint subsets from $\mathcal{V}$, where $\mathcal{V}_a$ refers to all the anomalous nodes  and $\mathcal{V}_n$ denotes all the normal nodes. Graph-based anomaly detection is to classify unlabeled nodes in $G$ into the normal or anomalous categories given the information of the graph structure $\mathbf{A}$, node features $\textbf{X}$, and partial node labels from $\mathcal{V}_a$ and $\mathcal{V}_n$.

\label{def:anomalyProblem}
\end{problem}

Usually, there are far more normal nodes than anomalous nodes, $\lvert \mathcal{V}_a \rvert \ll \lvert \mathcal{V}_n \rvert$, thus graph-based anomaly detection can be regarded as an extremely imbalanced binary node classification problem. The main difference is that anomaly detection focuses more on the unusual and deviated patterns in the dataset.
A summary of the symbols used in this article is presented in Table~\ref{tab:Notations}.

\subsection{Denoising Diffusion Probabilistic Models}
\label{ddpm}

Denoising diffusion probabilistic models (DDPM)~\cite{ho2020denoising,sohl2015deep} is a class of generative models that show superior performance in unconditional image generation, compared to GANs~\cite{xiao2022tackling,gao2021adversarial}. It learns a Markov Chain which gradually converts a simple distribution (e.g., isotropic Gaussian) into a data distribution. The generative process learns the reverse of the DDPM's forward (diffusion) process: a fixed Markov Chain that gradually adds noise to data. Here, each step in the forward process is a Gaussian translation:

\begin{equation}
\begin{split}
    q(\textbf{z}^t \lvert \textbf{z}^{t-1}):=N(\textbf{z}^t;\sqrt{1-\beta_t} \textbf{z}^{t-1},\beta_t\textbf{I}),
\end{split}
\label{equ:forwardProcess}
\end{equation}
where $\beta_1, ..., \beta_T$ is a fixed variance schedule rather than learned parameters ~\cite{ho2020denoising}. Eq.~\eqref{equ:forwardProcess} is a process finding $\textbf{z}^t$ by adding a small Gaussian noise to the latent variable $\textbf{z}^{t-1}$. Given clean data $\textbf{z}^0$, sampling of $\textbf{z}^t$ can be expressed in a closed form:
\begin{equation}
\begin{split}
    q(\textbf{z}^t|\textbf{z}^0):=N(\textbf{z}^t;\sqrt{\bar{\alpha}_t} \textbf{z}^0, (1-\bar{\alpha}_t)\textbf{I}),
\end{split}
\end{equation}
where $\alpha_t := 1-\beta_t $ and $\bar{\alpha}_t :=\prod_{s=1}^t \alpha_s$. Therefore, $\textbf{z}^t$ is expressed as a linear combination of $\textbf{z}^0$ and $\epsilon$:
\begin{equation}
\begin{split}
    \textbf{z}^t = \sqrt{\bar{\alpha}_t} \textbf{z}^0 + \sqrt{1-\bar{\alpha}_t}\epsilon,
\end{split}
\label{equ:forwordSampling}
\end{equation}
where $\epsilon \sim N(0,\textbf{I})$ has the same dimensionality as data $\textbf{z}^0$ and latent variables $\textbf{z}^1,...,\textbf{z}^T$.

Since the reverse of the forward process, $q(\textbf{z}^{t-1} \lvert \textbf{z}^t)$, is intractable, DDPM learns parameterized Gaussian transitions $p_\theta(\textbf{z}^{t-1} \lvert \textbf{z}^t)$. The generative (or reverse) process has the same functional form ~\cite{sohl2015deep} as the forward process, and it is expressed as a Gaussian transition with learned mean and fixed variance ~\cite{ho2020denoising}:
\begin{equation}
\begin{split}
    p_\theta(\textbf{z}^{t-1} \lvert \textbf{z}^t) = N(\textbf{z}^{t-1};\mu_\theta(\textbf{z}^t, t),\sigma_t^2\textbf{I} ).
\end{split}
\label{equ:reverseCond}
\end{equation}
Further, by decomposing $\mu_\theta$ into a linear combination of $\textbf{z}^t$ and the noise approximator $\epsilon_\theta$, the generative process is expressed as:

\begin{equation}
\begin{split}
    \textbf{z}^{t-1}= \frac{1}{\sqrt{\alpha_t}}(\textbf{z}^{t}-\frac{1-\alpha_t}{\sqrt{1-\bar{\alpha}_t}}\epsilon_\theta (\textbf{z}^t, t))+\sigma_t \epsilon,
\end{split}
\label{equ:reverseProcess}
\end{equation}
which suggests that each generation step is stochastic.
Here $\epsilon_\theta$ represents a neural network with the same input and output dimensions and the noise predicted by the neural network $\epsilon_\theta$  in each step is used for the denoising process in Eq.~\eqref{equ:reverseProcess}.

\subsection{Graph Pointer Neural Network}
\label{sec:pointer}

The Graph Pointer Neural Network (GPNN)~\cite{yang2022graph} aims to select the most relevant and valuable neighboring nodes, which constructs a new sequence ranked by relations to the central node. GPNN takes a node sequence consisting of the neighbors of a given central node as input, and outputs an ordered sequence according to the relationship with the central node. This model utilizes a sequence-to-sequence architecture based on LSTMs for the pointer network to order the sequence.

Specifically, GPNN is composed of an \textit{Encoder} and a \textit{Decoder}.
The encoder generates hidden states for the input sequence. At each time step $i$, the feature vector of node $\hat{\mathbf{x}}_i$ is fed into the encoder, the hidden state is denoted as:
\begin{equation} \begin{aligned}
\mathbf{e}_i = \tanh(\mathbf{W}[\mathbf{e}_{i-1},\hat{\mathbf{x}}_i]),
\end{aligned} \end{equation}
where $\mathbf{e}_0$ is initialed to zero vector. After $L$ time steps, GPNN obtains $L$ hidden states of input sequence and combines them into a content vector $\mathbf{E} = \{\mathbf{e}_1, \mathbf{e}_2,\dots, \mathbf{e}_L\}$ that records the information of entire sequence of $L$ nodes.

The decoder then selects nodes with attention scores among the $L$ nodes. At each output time $i$, the hidden state of decoder $d_i$ is obtained by:
\begin{equation}
\begin{aligned}
\mathbf{d}_i = \tanh(\mathbf{W}[\mathbf{d}_{i-1}, \hat{\mathbf{x}}_{c_{i-1}}]),
\end{aligned} \end{equation}
where $\mathbf{d}_0 = \mathbf{e}_L$ is the output hidden state from the encoder and $c_{i-1}$ is the index of selected node at time step $i-1$. GPNN computes the attention vectors over the input sequence as following:
\begin{equation} \begin{aligned}
u^i_j =\beta^T \tanh(\mathbf{W}_1 \mathbf{e}_j+ \mathbf{W}_2 \mathbf{d}_i), \qquad j \in (1,2,\ldots,L),
\end{aligned} \end{equation}
\begin{equation} \begin{aligned}
p(c_i|c_1,c_2,...c_{i-1},s) = \text{softmax}(\mathbf{u}_i),
\end{aligned} \end{equation}
where $\beta^T, \mathbf{W}_1,$ and $\mathbf{W}_2$ are learnable parameters of the decoder model and $\text{softmax}$ normalizes the vector $\mathbf{u}_i$ (of length $L$) to be an output distribution over the $L$ nodes of the input sequence. With the probability distribution $\mathbf{u}_i$, GPNN uses $u^i_j$ as pointers to select the $i$-th node of output sequence, until all top-$k$ nodes are selected step by step. After $k$ output time steps, GPNN obtains the sequence of top-$k$ relevant nodes as:
$\{\hat{\mathbf{x}}_{c1},\hat{\mathbf{x}}_{c2},\ldots,\hat{\mathbf{x}}_{ck} \} 
(\in \mathbb{R}^{k \times d}) $,
which is ranked with the output order.

\begin{figure}[t]
    \centering
    \includegraphics[width=\linewidth]{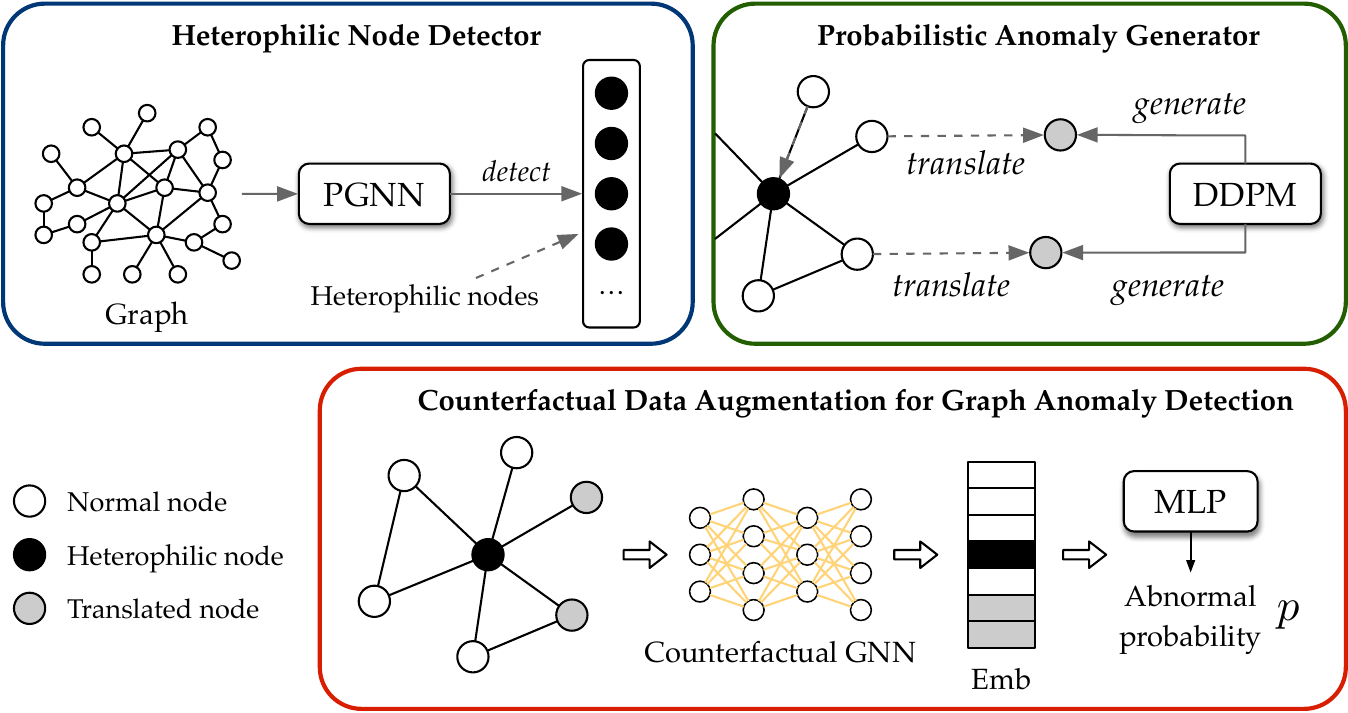}

    \caption{Sketch of CAGAD. Heterophilic nodes are defined as most of their neighbors have different properties or labels from themselves.}
    \label{fig:teasear}
\end{figure}

\section{CAGAD: Detailed Methodology}
\label{sec:method}

We now discuss the CAGAD framework in detail. We firstly present the heterophilic node detector that identifies nodes with heterophily-dominant neighbors in an unsupervised manner.
Then we introduce the DDPM-based anomaly generator, which aims to translate the source embedding into the ones with anomalous labels. 
Lastly, we incorporate the generator into GNNs to manufacture counterfactual augmented representations by aggregating generated unseen neighbor embeddings. These augmented representations will be used for enhancing anomaly detection. A sketch of CAGAD is shown in \figurename~\ref{fig:teasear}.

\subsection{Heterophilic Node Detector}

We first present the heterophilic node detector to identify the nodes whose neighbors are heterophily-dominant without requiring labeled data. The graph pointer neural network (PGNN)~\cite{yang2022graph} is used to build the detector,
which
calculates the heterophily-degree for heterophilic node detection via attention scores.
Here we aim to distinguish crucial information from distant nodes while filtering out irrelevant or noisy ones in the nearest neighborhood.
To this end, we leverage a pointer network to compute attention vectors (scores) and then select the most relevant nodes from multi-hop neighborhoods according to these scores.
Since the attention scores can denote the relevant relationship with the target node~\cite{yang2022graph},
we use them to identify heterophilic nodes: a node is regarded as the heterophilic one if most of its neighbors have lower attention scores than a given threshold, say $\eta$.
The optimal threshold $\eta$ can be determined by the existing training data.

Specifically,
we first construct a node sequence for each target node that contains neighbors within a given hop, such as one-hop.
As the number of neighbors varies from node to node, we set a fixed maximum length
$L$ of the sequence and stop sampling when we meet this limitation.
Note that truncating neighbors will not impact detection results since the heterophilic nodes are determined by the ratios instead of absolute node number.
When some nodes have a small number of neighbors, we duplicate one neighbor repeatedly until reaching $L$ and, after obtaining the attention scores, the duplicated ones will be deleted.
Then, we apply a GCN layer~\cite{kipf2016semi} to compute local embeddings, which captures the local information of each node. With the input nodes feature $\textbf{X}$, the output embedding is denoted as:
\begin{equation} \begin{aligned}
    \hat{\textbf{X}} = GCN(\textbf{X}),
\label{equ:gcnX}
\end{aligned} \end{equation}
where we
embed the feature vectors into  hidden representations.

Next, we use a pointer network to calculate the attention scores for the node sequence.
A sequence-to-sequence architecture based on LSTMs is adopted for the pointer network, 
composing 
an Encoder and a Decoder.
For the target node $v_i$ and one of its neighbors $v_j$, the attention score $u^i_j$ is computed as follows: 
\begin{equation} \begin{aligned}
    u^i_j = \beta^T \tanh (\mathbf{W}_1 \mathbf{e}_j + \mathbf{W}_2 \mathbf{d}_i),
\label{equ:attScore}
\end{aligned} \end{equation}
where $\beta^T$, $\mathbf{W}_1$ and $\mathbf{W}_2$ are learnable parameters of the model, and $\mathbf{e}_j$ and $\mathbf{d}_i$ are the the hidden states of the encoder and decoder, respectively (see Section~\ref{sec:pointer} for details).

Then we are able to compute the desired heterophily degree $h_d$ to determine whether the target node is a heterophilic one via:
\begin{equation} \begin{aligned}
    h_d = |\mathcal{V}_i^\eta| /  |\mathcal{V}_i^S|,
\label{equ:attScoreTh}
\end{aligned} \end{equation}
where $\mathcal{V}_i^\eta = \{v: u^i_j < \eta\}$ is the set of the neighbors whose attention scores are less than the threshold $\eta$, and $\mathcal{V}_i^S$ is the set of $v_i$'s neighbors which are included in the sequence. Obviously, $h_d$ represents the percentage of heterophilic neighbors among all neighbors of the target node. For example, $h_d$=60\% means that 60\% of its neighbors have the different class labels from it.
If heterophily degree 
$h_d$ of a node is greater 
than a given value $\alpha$, that node 
is considered 
heterophilic. 

Based on the above, 
we can extract all the heterophilic nodes of a graph.
For each of them, we select a given fraction of neighbors with lower attention scores and translate them into anomalous nodes. These generated anomalous neighbors are used to replace the original ones for neighborhood aggregation.
If the heterophilic target node is anomalous, most of its neighbors should be normal.
Some normal neighbors will be replaced with generated anomalous ones. This replacement can make the representations more distinguishable.
If the heterophilic target node is normal, most of its neighbors should be anomalous. Again, some anomalous neighbors will be replaced with generated anomalous ones.
This replacement will hardly influence the target node, since the number of anomalous neighbors keeps unchanged.
Thus, manipulating heterophilic node neighbors will not impact normal node representations but can benefit anomaly representations to boost performance.

\subsection{DDPM-Based Anomaly Generator}
\label{sec:CoreSubgraph}

Having identified nodes with heterophily-dominant neighbors, we now 
introduce how the node embeddings are translated into anomalous ones.
Inspired by the superiority of diffusion models in image generation and translation~\cite{meng2022sdedit,sinha2021d2c,xiao2023imputation}, we design a graph-specific probabilistic diffusion model as the anomaly embedding generator. This model can iteratively extract distinct features from anomalies and then exert them on the generative process to manufacture anomalous nodes.
This generator takes a source embedding $\mathbf{z}_\text{src}$ without label information and a reference embedding $\mathbf{z}_\text{ref}$ with the anomalous label as inputs, and translates the source embedding into anomalous one.
The source embedding can be either normal or anomalous one.
Here, the reference embedding can be obtained by using the aforementioned GCN layer~\cite{kipf2016semi} or a trained GNN-based classifier.

\begin{figure}[t]
\centering
\includegraphics[width=0.425\textwidth]{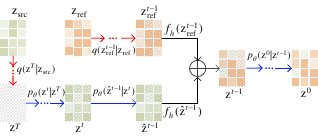}

\caption{Anomalous generator $G_\text{ano}$: red arrows $\rightarrow$ indicate forward diffusion, blue ones $\rightarrow$ refer to the reverse diffusion; $\oplus$ is the concatenation operation.}
\label{fig:generatorAno}
\end{figure}

\figurename~\ref{fig:generatorAno} presents the framework of the anomaly embedding generator $G_\text{ano}$.
The generator first adopts the forward process to add noise into $\textbf{z}_\text{src}$ to form a prior $\textbf{z}^T$.
Then the prior $\textbf{z}^T$ is fed into the reverse diffusion process to generate a clean embedding through gradual denoising,
i.e., $\textbf{z}^T \rightarrow \textbf{z}^t \rightarrow \hat{\textbf{z}}^{t-1} \rightarrow \textbf{z}^{t-1} \rightarrow \textbf{z}^0$.
During this denoising process, the distinct features of anomalous nodes are extracted from $\textbf{z}_\text{ref}$ and iteratively injected into the latent variable $\hat{\textbf{z}}^{t-1}$.
In this way, the generated embedding $\mathbf{z}^0$ not only has the anomalous label of the reference embedding $\mathbf{z}_\text{ref}$, but also keeps the characteristics of the source embedding $\mathbf{z}_\text{src}$.

Concretely, we first utilize the forward process $q(\textbf{z}^t|\textbf{z}^0)$ to generate a prior $\textbf{z}^T$ by adding noise:
\begin{equation} \begin{aligned}
    \textbf{z}^T = \sqrt{\bar{\alpha}_T} \textbf{z}_\text{src} + \sqrt{1-\bar{\alpha}_T}\epsilon,
\label{equ:priorZT}
\end{aligned} \end{equation}
where $\bar{\alpha}_T =\prod_{t=1}^T (1-\beta_t)$ and $\beta_1, ..., \beta_T$ is a fixed variance schedule~\cite{ho2020denoising}.
Then, we feed the prior $\textbf{z}^T$ into the reverse diffusion process and exert the condition $c_\text{ano}$ on this reverse process to generate an embedding with the anomalous label.
We approximate the Markov transition under the condition $c_\text{ano}$ as follows:
\begin{equation} \begin{aligned}
    p_\theta(\textbf{z}^{t-1}|\textbf{z}^{t},c_\text{ano}) \approx p_\theta \Big(\textbf{z}^{t-1}|\textbf{z}^{t},f_h(\textbf{z}^{t-1})=f_h(\textbf{z}^{t-1}_\text{ref})\Big),
\label{equ:OppCondition}
\end{aligned} \end{equation}
where $\textbf{z}^{t-1}_\text{ref}$ is sampled by Eq.~\eqref{equ:forwordSampling} and $f_h(\cdot)$ is a high-pass filter.

In each Markov transition, Eq.~\eqref{equ:OppCondition} tries to incorporate the high-frequency information extracted from $\textbf{z}_\text{ref}$ by $f_h(\cdot)$ into the generated embedding.
Since anomalies tend to have different features from their neighbors (high-frequency information), and normal nodes tend to share common features with their normal neighbors (low-frequency information)~\cite{chai2022can,tang2022rethinking}, high-frequency information can be regarded as the distinct features of the anomalies.
Hence, we utilize the high-pass filter $f_h(\cdot)$ to extract it 
from $\textbf{z}_\text{ref}^{t-1}$ (corrupted $\textbf{z}_\text{ref}$), and inject this information into the generated embedding. As a result, the generated embedding by Eq.~\eqref{equ:OppCondition} will incline to possess the anomalous information (label).
Since the transition of Eq.~\eqref{equ:OppCondition} starts with the prior $\textbf{z}^T$ 
(derived from $\textbf{z}_\text{src}$), the generated embedding also has the characteristics of source embedding $\textbf{z}_\text{src}$.

According to Eq.~\eqref{equ:OppCondition}, in each transition from $\textbf{z}_t$ to $\textbf{z}_{t-1}$, the distinct features of anomalies $f_h(\textbf{z}_\text{ref}^{t-1})$ are extracted from the corrupted reference embedding $\textbf{z}_\text{ref}^{t-1}$ and then injected into the latent variable.
To this end, we first adopt the forward process (Eq.~\eqref{equ:forwordSampling}) to compute $\textbf{z}_\text{ref}^{t-1}$ from $\textbf{z}_\text{ref}$, and then adopt the reverse process (Eq.~\eqref{equ:reverseCond}) to calculate latent variable $\hat{\textbf{z}}^{t-1}$ from $\textbf{z}^t$.
Since high-filter operation $f_h(\cdot)$ maintains the dimensionality of the input, we refine the latent variables by matching $f_h(\hat{\textbf{z}}^{t-1})$ of $\hat{\textbf{z}}^{t-1}$ with that of $\textbf{z}_\text{ref}^{t-1}$ as follows:
\begin{equation}
\begin{split}
    &\textbf{z}_\text{ref}^{t-1} \sim q(\textbf{z}_\text{ref}^{t-1} \lvert \textbf{z}_\text{ref}),
    \\
    &\hat{\textbf{z}}^{t-1} \sim p_\theta(\hat{\textbf{z}}^{t-1} \lvert \textbf{z}^t),
    \\
    \textbf{z}^{t-1}= &\hat{\textbf{z}}^{t-1} + \gamma \left(f_h(\textbf{z}_\text{ref}^{t-1})-f_h(\hat{\textbf{z}}^{t-1})\right),
\end{split}
\label{equ:OppFinal}
\end{equation}
where $\gamma$ is a weight parameter to adjust the importance of the high-frequency information.
The matching operation by Eq.~\eqref{equ:OppFinal} ensures the condition $c_\text{ano}$ in Eq.~\eqref{equ:OppCondition}, which further enables the conditional generation based on DDPM.
In this way, through injecting distinct features of anomalies into the latent variable in the generative process, the generated embedding can possess the anomalous label of $\textbf{z}_\text{ref}$.
At the same time, the input of the generative process is the prior derived from $\textbf{z}_\text{src}$, thus the generated embedding still has characteristics of $\textbf{z}_\text{src}$.


\subsection{Counterfactual Graph Neural Network}

\begin{figure}[t]
        \begin{center}
        \subfloat[Original aggregation]{
        \label{fig:GNNorignal}\includegraphics[width=0.19\textwidth]{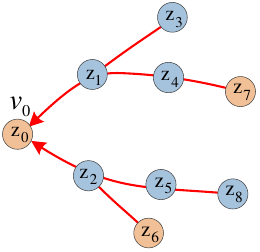}}
        \subfloat[Aggregation via counterfactuals]{
        \label{fig:GnnCounter}\includegraphics[width=0.270\textwidth]{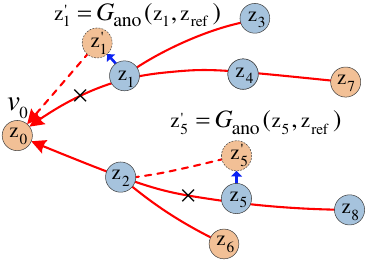}}

        \caption{Original GNN vs. Counterfactual GNN.}
        \label{fig:conterGNN}
        \end{center}
\end{figure}

We here incorporate the anomaly embedding generator $G_\text{ano}$ into a graph attention network to build the counterfactual GNN, which can generate effective augmented data by involving generated anomaly embeddings into the neighborhood aggregation process.
Figure~\ref{fig:conterGNN} presents the neighborhood aggregation process.
For the original GNN in Figure~\ref{fig:GNNorignal}, the representation of target node $v_0$ is obtained by aggregating the embeddings of its neighbors directly. For the 
counterfactual GNN in Figure~\ref{fig:GnnCounter}, 
before neighborhood aggregation, we first select the node with heterophily-dominant neighbors (such as $v_0$), and translate a part of its neighbors (e.g., 
$v_1$ and $v_5$) into anomalous ones (e.g., 
$v_1^{'}$ and $v_5^{'}$) by using 
$G_\text{ano}$.
Then the generated nodes are used to replace the original neighbors for GNN neighborhood aggregation to compute representation $\textbf{z}_0$ of target node $v_0$.

A 
representation 
computed by the counterfactual GNN is regarded as counterfactual data because its generation process involves 
unseen nodes.
Moreover, the counterfactual representation will have the same label as the one calculated by the original GNN, as 
changing a few edges/neighbors 
will not greatly impact the node identification (label) information \cite{zhu2021graph,zheng2021generative}.
Hence, this counterfactual GNN can be 
applied 
to test data for obtaining better node representations.

We use the graph attention networks (GATs) \cite{velivckovic2018graph} as the backbone of the counterfactual GNN
, which introduces the masked attention mechanism to represent the importance of different adjacent nodes.
Formally, in each layer $l-1$, node $v_i$ integrates the features of neighboring nodes to obtain representations of layer $l$ via:
\begin{equation} \begin{aligned}
\textbf{h}^{(l)}_i = \sigma \left( \sum\nolimits_{j \in \mathcal{V}_i \cup \mathcal{V}_i^{'} \cup \{v_i\} } a_{ij}
\textbf{W} \cdot \textbf{h}^{(l-1)}_j   \right),
\label{equ:GnnFinalRepr}
\end{aligned} \end{equation}
where $\sigma$ refers to a nonlinear activation function (e.g., ReLU), $\mathcal{V}_i$ is the set of neighbors for $v_i$ (excluding the neighbors 
selected as the source ones), $\mathcal{V}_i^{'}$ is the set of neighbors corresponding to the embeddings generated by $G_\text{ano}$, and $a_{ij}$ 
is the attention coefficient between node $v_i$ and node $v_j$, which can be computed as:
\begin{equation} \begin{aligned}
a_{ij} = \frac{\text{exp}(\sigma (\text{a}^\text{T} [\textbf{W} \textbf{h}_i^{(l)} \oplus \textbf{W} \textbf{h}_j^{(l)}]  ) )}{\sum\nolimits_{k \in \mathcal{V}_i \cup \mathcal{V}_i^{'} \cup \{v_i\}} \text{exp}( \sigma(a^\text{T}[\textbf{W} \textbf{h}_i^{(l)} \oplus \textbf{W} \textbf{h}_k^{(l)}] ))},
\label{equ:attCoefficient}
\end{aligned} \end{equation}
where $\oplus$ is the concatenation operation and attention vector $a$ is a trainable weight vector that assigns importance to different neighbors of node $v_i$, allowing the model to highlight the features of the important neighboring node that is more task-relevant.

To incorporate high-order neighborhood, multiple layers are adopted to build the graph attentive encoder:
\begin{equation} \begin{aligned}
\textbf{h}^{(1)}_i &= \sigma \left( \sum\nolimits_{j \in \mathcal{V}_i \cup \mathcal{V}_i^{'} \cup \{v_i\}} a_{ij}^{(1)}
\textbf{W}^{(1)} \cdot \textbf{x}_j   \right),
\\& ......\\
\textbf{z}_i &= \sigma \left( \sum\nolimits_{j \in \mathcal{V}_i \cup \mathcal{V}_i^{'} \cup \{v_i\}} a_{ij}^{(L)}
\textbf{W}^{(L)} \cdot \textbf{h}^{(l-1)}_j   \right),
\label{equ:multipleGAT}
\end{aligned} \end{equation}
where $\textbf{z}_i$ is the latent representation of node $v_i$. 
In this way, the graph attentive encoder is able to map the learned node representations by capturing the nonlinearity of topological structure and node attributes.

The aggregated representations $\textbf{z}_i$ are then fed into another MLP with a Sigmoid function to compute the abnormal probability $p_i$. The weighted cross-entropy loss is then used for the model training:

\begin{equation} \begin{aligned}
\mathcal{L} = \sum_i \left( \varphi y_i \log(p_i) + (1-y_i) \log(1-p_i) \right),
\label{equ:crossLoss}
\end{aligned} \end{equation}
where $\varphi$ is the proportion of anomaly labels ($y_i=1$) to normal labels ($y_i=0$).

This model can improve the detection performance by considering augmented counterfactual representations because the counterfactual representations of anomalies are enhanced by involving information of generated anomalous neighbors.
The proposed data augmentation method scarcely requires labeled data since its main components -- the heterophilic node detector and DDPM-based generator -- are unsupervised.
At the same time, this method can also be used to enhance node representations of test data, because it requires no labels for the test data and the computed counterfactual representations have the same labels as the source ones.

\subsection{Discussions}
\label{sec:Disscusion}

In this section, we show the connection between our model and GNNs as well as causal representation learning on graphs.

\subsubsection{Relation to GNNs}
\label{sec:RelationGNNs}

GNN has been a mainstream technique for graph anomaly detection due to its ability to learn expressive node representations by aggregating node's local neighborhoods~\cite{xu2019powerful,xu2022ccgl}. One of the key components in CAGAD is the counterfactual GNN which manipulates the neighborhood aggregation to produce counterfactual node representations. In the following, we present a detailed comparison between the vanilla GNN and the proposed counterfactual GNN.

\noindent\textbf{The original GNN}.
Given the adjacency matrix $\textbf{A}$ indicating connections between nodes in $\mathcal{V}$ and the node attributes $\textbf{X}$,
the original GNN computes the node representations by aggregating features from the local neighborhoods iteratively:
\begin{equation}
\begin{aligned}
\textbf{H}^1 &= \text{GNN}^1(\textbf{A},\textbf{X}), \\
\textbf{H}^2 &= \text{GNN}^2(\textbf{A},\textbf{H}^1), \\
             &\quad \dots, \quad \\
 \textbf{Z}  &= \text{GNN}^L (\textbf{A}, \textbf{H}^{L-1}),
\label{equ:gnnOriginal}
\end{aligned}
\end{equation}
where $\text{GNN}^L$ is the last aggregation layer and $\textbf{Z}=\left[ \textbf{z}_{1}, ..., \textbf{z}_n\right ]$ is the final learned node representations from the GNN.
The learned representations $\textbf{Z}$ can 
be adopted to identify anomalies from normal nodes by classification-based models or reconstruction-error-based methods.

\noindent\textbf{The counterfactual GNN}.
In the proposed model, according to the detection results obtained by the heterophilic node detector (whether nodes are the heterophilic ones),
node representations $\textbf{Z}=\left[ \textbf{z}_{1}^o, ..., \textbf{z}_{m}^o, \textbf{z}_{m+1}^h, ..., \textbf{z}_n^h \right ]$ are split into two groups: one for non-heterophilic node representations $\textbf{Z}^o = \left[ \textbf{z}_{1}^o, ..., \textbf{z}_{m}^o \right ]$ and another for heterophilic node representations $\textbf{Z}^h = \left[ \textbf{z}_{m+1}^h, ..., \textbf{z}_n^h \right ]$.
For the non-heterophilic nodes, their representations $\textbf{Z}^o$ are still computed by using the original GNN (Eq.~\eqref{equ:gnnOriginal}).

On the other hand, for the heterophilic nodes, their representations $\textbf{Z}^h$ are learned by the counterfactual GNN, where a part of neighbors will be translated into the new ones with anomalous labels. The aggregating process is expressed as:
\begin{equation} \begin{aligned}
\textbf{H}^1 &= \text{GNN}^1(\textbf{A},\textbf{X}), \\
\widehat{\textbf{H}}^1_s &= G_{\text{ano}}(\textbf{H}^1_s, \textbf{Z}_{\text{ref}}), \\
\textbf{H}^2 &= \text{GNN}^2(\textbf{A}, (\textbf{H}^1 - \textbf{H}^1_s) \cup \widehat{\textbf{H}}^1_s), \\
             &\quad \dots, \quad \\
 \textbf{Z}^h  &= \text{GNN}^L (\textbf{A}, \textbf{H}^{L-1}),
\label{equ:gnnCounter}
\end{aligned} \end{equation}
where $\textbf{H}^1_s$ is the embedding set of selected neighbors to be translated into anomalous ones.
The counterfactual GNN still takes the adjacency matrix $\textbf{A}$ and the node attributes $\textbf{X}$ as inputs. But during iterative aggregation, a part of the neighbor embeddings of the heterophilic nodes are translated into the ones with anomalous labels by our designed generator $G_\text{ano}$, and these generated embeddings are involved for learning node representations.
By manipulating the neighborhood aggregation, we can manufacture counterfactual representations that are more distinguishable from normal node representations, and subsequently improve the detection performance.

\subsubsection{Relation to Causal Representation Learning on Graph}
\label{sec:RelationCausal}

Causal representation learning aims to learn a representation exposing the causal relations that are invariant under different interventions. Generating counterfactual data is an effective way to remove spurious correlations and help learn better representations and further improve the classification performance.
Existing data augmentation methods for graph data mainly focus on manipulating edges or node attributes, and then they utilize GNNs to generate counterfactual augmented data.
Assuming that $\widehat{\textbf{A}}$ and $\widehat{\textbf{X}}$ denote the manipulated adjacency matrix and attribute matrix individually, general augmentation methods adopt the plain GNN to produce augmented data by considering the manipulated adjacency matrix: $\textbf{Z} = \text{GNN}(\widehat{\textbf{A}}, \textbf{X})$, or the manipulated attribute matrix: $\textbf{Z} = \text{GNN}(\textbf{A}, \widehat{\textbf{X}})$, or both: $\textbf{Z} = \text{GNN}(\widehat{\textbf{A}}, \widehat{\textbf{X}})$.

By contrast, our model CAGAD differentiates them in two aspects:
(\textit{i}) Instead of simply changing edges and attributes, CAGAD manipulates the neighborhood aggregation process to produce counterfactual augmented data, which can generate more flexible augmented data.
(\textit{ii}) Unlike these methods which principally focus on enhancing training data, our model can be applied to test data in boosting representation of anomalies.

\section{Experimental Evaluation}
\label{sec:experiment}


In this section, we conduct empirical evaluations to demonstrate the efficacy of our approach concerning anomaly detection performance, and the impact of the translated neighbor number and the counterfactual augmentation technique.


\begin{table}[t]
\small
\caption{Statistics of Datasets
} 
\label{table:datasets}
\centering
\begin{tabular}{lrrrrrr}
\toprule
  Datasets       & \# node  & \# edge & \# feature & \ anomaly (\%)  \\
\midrule
  PubMed         &19,717     & 44,338       & 500	  & 20.81\%  \\
  T-finance    &39,357     & 21,222,543   & 10      & 4.58\% \\
  Amazon         &11,944     & 4,398,392    & 25	  & 6.87\%  \\
  YelpChi        &45,954     & 3,846,979    & 32	  & 14.53\%  \\
\bottomrule
\end{tabular}
\end{table}

\begin{table*}[t]
\small
\setlength{\tabcolsep}{5pt}
\centering
\caption{Performance comparison between CAGAD and baselines. We bold the best result and underline the second.}
\label{tab:detectResults}%
\begin{tabular}{lcccccccccccc}
\toprule
             \textbf{Method}               & \multicolumn{3}{c}{\textbf{PubMed}}    & \multicolumn{3}{c}{\textbf{T-Finance}}      & \multicolumn{3}{c}{\textbf{Amazon}}   & \multicolumn{3}{c}{\textbf{YelpChi}}    \\
\cmidrule(lr){2-4} \cmidrule(lr){5-7} \cmidrule(l){8-10} \cmidrule(l){11-13}
           & \footnotesize{F1}  & \footnotesize{AUC-ROC}  & \footnotesize{AUC-PR}
           & \footnotesize{F1}  & \footnotesize{AUC-ROC}  & \footnotesize{AUC-PR}
           & \footnotesize{F1}  & \footnotesize{AUC-ROC}  & \footnotesize{AUC-PR}
           & \footnotesize{F1}  & \footnotesize{AUC-ROC}  & \footnotesize{AUC-PR}
            \\
\midrule

        GAT & 55.15  & 53.24  & 46.24  & 53.15  & 52.04  & 43.10  & 60.84  & 73.45  & 68.97  & 50.27  & 50.95  & 24.06  \\
        GIN & 59.25  & 69.48  & 60.35  & 58.25  & 68.86  & 57.03  & 68.69  & 78.83  & 70.22  & 57.57  & 64.73  & 30.57  \\
        GraphSAGE & 59.63  & 66.35  & 57.63  & 59.03  & 66.35  & 54.95  & 70.78  & 75.37  & 70.10  & 58.41  & 67.58  & 31.92  \\
        GWNN & 71.64  & 87.68  & 76.16  & 70.64  & 86.68  & 71.79  & 87.01  & 85.37  & 79.40  & 59.10  & 67.16  & 31.72  \\
        \midrule
        LA-GNN & 73.15  & 89.05  & 77.35  & 72.15  & 88.05  & 72.92  & 65.15  & 84.15  & 78.26  & 57.25  & 68.05  & 34.57  \\
        GCA & 74.54  & 90.28  & 78.89  & 73.54  & 88.28  & 73.11  & 66.54  & 86.62  & 80.56  & 59.34  & 68.28  & 34.71  \\
        \midrule
        GraphConsis & 73.39  & 90.56  & 78.66  & 71.73  & 90.28  & 74.75  & 68.59  & 74.11  & 68.92  & 56.79  & 66.41  & 37.46  \\
        CAREGNN & 75.47  & 90.83  & 75.98  & 73.32  & 90.50  & 71.95  & 68.78  & 88.69  & 80.47  & 62.18  & 75.07  & 39.26  \\
        PC-GNN & 63.06  & 91.37  & 79.36  & 62.06  & 90.76  & 75.17  & 79.86  & \underline{90.40}  & \underline{84.07}  & 59.82  & 75.47  & 40.58  \\
        BWGNN & \underline{84.49}  & \underline{93.51}  & \underline{81.22}  & \underline{84.89}  & \underline{91.15}  & \underline{75.49}  & \underline{90.92}  & 89.45  & 83.19  & \underline{67.02}  & \underline{76.95}  & \underline{43.41}  \\
        \midrule
        \textbf{CAGAD} & \textbf{86.61}  & \textbf{95.52}  & \textbf{84.39}  & \textbf{87.65}  & \textbf{93.34}  & \textbf{76.70}  & \textbf{92.30}  & \textbf{92.43}  & \textbf{85.23}  & \textbf{68.44}  & \textbf{78.67}  & \textbf{44.78} \\

\bottomrule
    \end{tabular}
\end{table*}

\subsection{Experimental Settings}

\noindent\textbf{Datasets.}
We perform experiments on four datasets introduced in Table 1.
\textit{PubMed}~\cite{2008collective} is a citation network of biomedical science, where the nodes and edges denote the scientific publication and citation connections between publications.
\textit{T-Finance}~\cite{tang2022rethinking} is a transaction network.
The nodes are unique anonymized accounts with 10-dimension features related to registration days, logging activities and interaction frequency. The edges in the graph are transaction records of accounts.
\textit{Amazon}~\cite{mcauley2013amateurs} is a review network, where nodes represent users and the edges indicate that two users have reviewed or rated the same product.
\textit{YelpChi}~\cite{rayana2015collective} is a review network collected from yelp.com, where nodes represent reviewers and links suggest two reviewers have commented the same product.

\noindent\textbf{Baselines.}
We compare the proposed framework with three types of baselines with different techniques, including general GNN models~\cite{velivckovic2018graph,xu2019powerful,hamilton2017inductive,xu2019graph}, graph data augmentation approaches~\cite{liu2022local,zhu2021graph},
and state-of-the-art methods for graph-based anomaly detection~\cite{liu2020alleviating,dou2020enhancing,liu2021pick,tang2022rethinking}.
\emph{GAT}~\cite{velivckovic2018graph} is a graph attention network that employs the attention mechanism for neighbor aggregation. \emph{GIN}~\cite{xu2019powerful} is  a GNN model connecting to the Weisfeiler-Lehman graph isomorphism test. \emph{GraphSAGE}~\cite{hamilton2017inductive} is a GNN model based on a fixed sample number of the neighboring nodes. \emph{GWNN}~\cite{xu2019graph} is a graph wavelet neural network using heat kernels to generate wavelet transforms.
\emph{LA-GNN}~\cite{liu2022local} is an efficient data augmentation strategy to generate
more features in the local neighborhood to enhance the expressive power of GNNs.
\emph{GCA}~\cite{zhu2021graph} is a contrastive framework that performs data augmentation on both topology and attribute levels to enhance node representations.
\emph{GraphConsis}~\cite{liu2020alleviating} is a heterogeneous GNN that tackles context, feature and relation inconsistency problems in graph anomaly detection. \emph{CAREGNN}~\cite{dou2020enhancing} is a camouflage-resistant GNN that enhances the aggregation process with the designed modules against camouflages. \emph{PC-GNN}~\cite{liu2021pick} is a GNN-based imbalanced learning method to solve the class imbalance problem in graph-based fraud detection via resampling. \emph{BWGNN}~\cite{tang2022rethinking} is a beta wavelet graph neural network, which can better capture anomaly information on a graph.

\noindent\textbf{Experiment Setup.}
We use 1,000 diffusion steps for the DDPM-based generators considering both efficiency and effectiveness.
For each heterophilic node, we empirically  select 70\% of its neighbors as the source nodes and translate their embeddings into anomalous embeddings.
We set the length $L$ of the sequence to the average of one-hop neighbors.
We train the detection models for 100 epochs by Adam optimizer with a learning rate of 0.01, and save the model with the best Macro-F1 on validation data for testing.
We use 1\% labeled data (both for normal and anomalous nodes) as the training set, the remaining data are split as validation and test sets by a ratio of 1:2.
We use the following three metrics to present a comprehensive evaluation: \emph{Macro-F1} is the unweighted mean of the F1-score of two classes. \emph{AUC-ROC} is the area under the ROC Curve, referring to the probability that a randomly chosen anomaly receives a higher score than a randomly chosen normal object. \emph{AUC-PR} is the area under the curve of precision against recall at different thresholds.
The models are implemented based on PyTorch and DGL. We conduct all the experiments on Ubuntu 20.04 with an Intel Core i9-12900K CPU, an NVIDIA GeForce RTX 3090 GPU, and 64GB memory.

\begin{figure*}[t]
\centering
\includegraphics[width=0.99\textwidth]{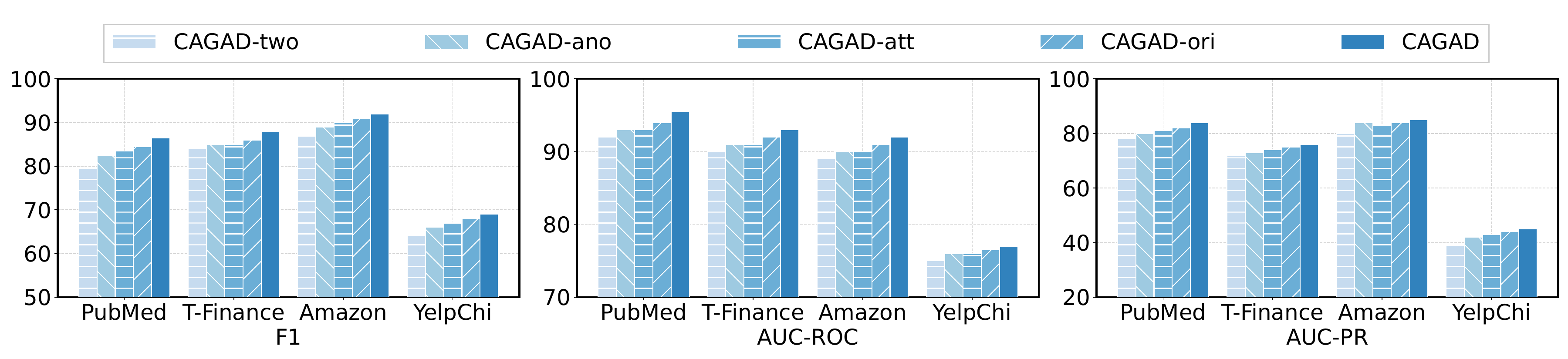}

\caption{Ablation study results on four datasets.}
\label{fig:Ablation}
\vspace{-4mm}
\end{figure*}

\subsection{Anomaly Detection Performance}
\label{sec:DetectionResults}

We report the anomaly detection performance of 
CAGAD and the baselines in Table~\ref{tab:detectResults} 
and we have the following observations: (1) For overall detection results, 
CAGAD 
yields uniformly better performance than all the baselines across four datasets.
In particular, CAGAD obtains on average 2.35\%, 2.53\% and 2.79\% improvements over the state-of-the-art, BWGNN, in terms of F1, AUC-ROC, and AUC-PR, respectively. 
(2) The improvements of CAGAD are more significant on T-Finance. We speculate this is because the anomalies in T-Finance are more sparse and the normal neighbors of anomalies may hinder the vanilla GNN from learning distinguishable representations.
Instead, CAGAD can generate augmented counterfactual data to acquire better anomaly representations, and further advocate the anomaly detection performance.
(3) The two graph data augmentation methods, LA-GNN and GCA, have better performances than the general GNN models (e.g., GraphSAGE and GWNN). 
These two models are even competitive with the models which are specifically designed for graph anomaly detection (e.g., CAREGNN and PC-GNN).
The reason is that only 1\% of nodes are adopted as labeled data for model training and data augmentation can effectively complement the limited training data. 
This result justifies our motivation that the proposed counterfactual data augmentation can enlarge the training data and be more effective than other augmentation-based or sampling-based GNN methods for anomaly detection.

Not surprisingly, the three anomaly detection baselines, GraphConsis, CAREGNN, and PC-GNN, outperform general GNN models. However, 
they ignore the 
problem that the anomalies with numerous benign neighbors might attenuate their suspiciousness under 
``vanilla'' neighborhood aggregation.
BWGNN, which is the best baseline, introduces the graph wavelet theory to remedy this issue, 
significantly improving 
the 
performance compared to others. 
However, by incorporating counterfactual augmented data, 
CAGAD can more effectively capture the anomaly information on the graph and 
outperforms BWGNN by a large margin.

\subsection{Ablation Study}

Here, we investigate the contributions of the essential components in CAGAD, i.e., the generated anomalous embeddings and the attention mechanism 
taking into account different components:
(1) \emph{CAGAD-two}, which removes the generated embeddings and the attention mechanism of the model;
(2) \emph{CAGAD-ano}, which removes the generated anomalous embeddings produced by $G_\text{ano}$ during neighborhood aggregation;
(3) \emph{CAGAD-att}, which removes the attention mechanism of neighborhood aggregation.
(4) \emph{CAGAD-ori}, which keeps the original embedding from being replaced by the generated one for neighborhood aggregation.

The 
results are shown 
in \figurename~\ref{fig:Ablation}. We summarize our observations as follows.
($i$) When removing the generated embeddings and attention mechanism, \emph{CAGAD-two} becomes a plain GNN and
performs the worst, 
suggesting 
that without the counterfactual augmented data, traditional neighborhood aggregation mechanisms cannot effectively capture anomalous information, resulting in poor performance. Nevertheless, our designed data augmentation method can address this issue and boost detection performance.
($ii$) \emph{CAGAD-ano} suffers from significant performance degradation due to the removal of the DDPM-based generator 
indicating 
that 
involving the generated anomalous embeddings for the neighborhood aggregation 
yields an obvious performance gain. The generated anomalous embeddings can make the representations of anomalies more distinguishable from normal nodes, and further improve detection performance.
($iii$) After eliminating the attention mechanism, the performance of \emph{CAGAD-att} 
drops significantly, which shows that the attention mechanism can enhance the detection performance. This is particularly true for our case as the counterfactual GNN requires adaptively adjusting the weights of the generated embeddings for obtaining better representations. Hence, it is also essential to include the attention mechanism in CAGAD.
($iv$) Involving the source embeddings for neighborhood aggregation lowers the performance slightly, which indicates that the designed replacement strategy is effective for anomaly detection.
By combining all the components 
CAGAD achieves the best performance.

\begin{figure*}[t]
\centering
\includegraphics[width=\textwidth]{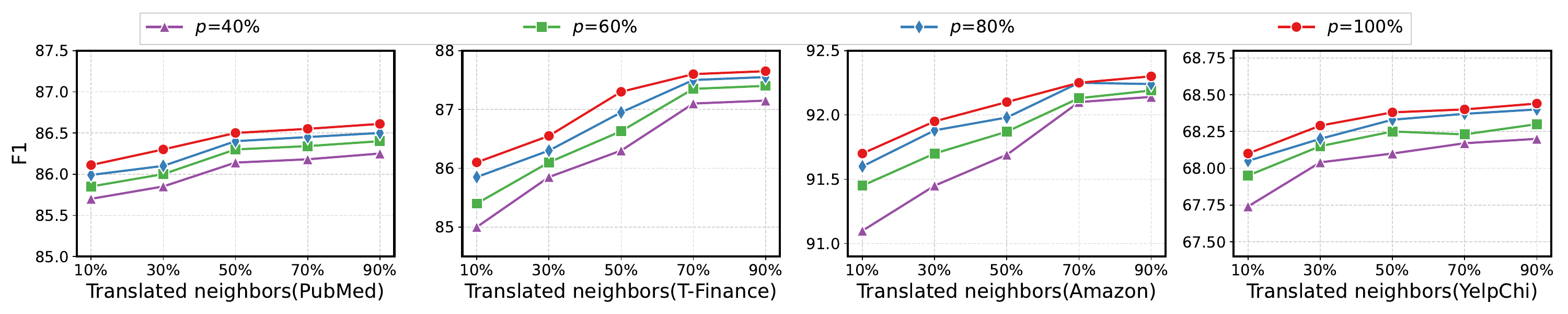}
\vspace{-3mm}
\caption{F1 performance with different numbers of translated neighbors.}
\label{fig:FigTranslatedF1}
\vspace{-3mm}
\end{figure*}

\begin{figure*}
\centering
\includegraphics[width=\textwidth]{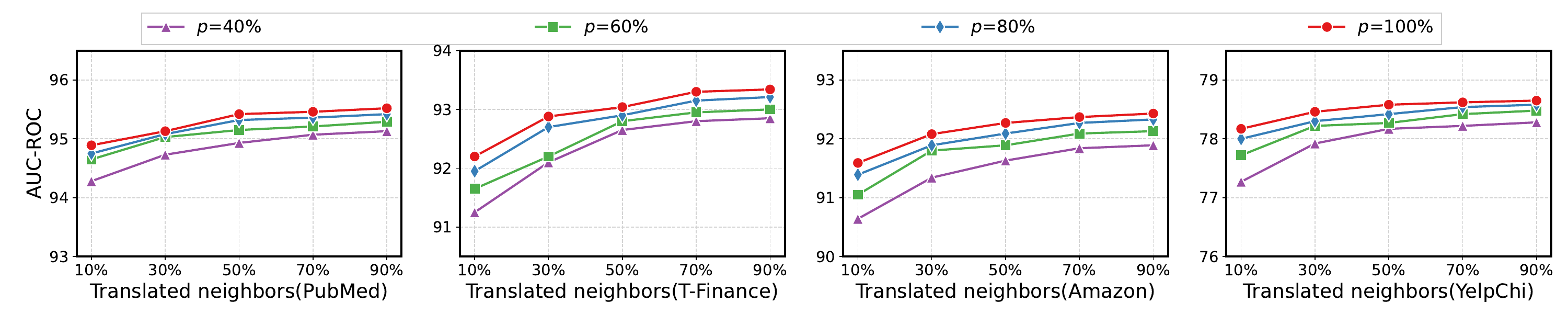}
\vspace{-3mm}
\caption{AUC-ROC performance with different numbers of translated neighbors.}
\label{fig:FigTranslatedAUC}
\vspace{-3mm}
\end{figure*}

\subsection{Influence of Translated Neighbors}
\label{sec:changeNeighbor}

The number of translated neighbors for heterophilic nodes is a critical hyper-parameter 
which 
affects the distinguishability of the learned representations and, subsequently, influences the anomaly detection performance.
We investigate the performance change by altering the number of translated neighbors.
In these experiments, for each obtained heterophilic node, we translate a different proportion of its neighbors into anomalous ones and 
observe the results when selecting $p=[40, 60, 80, 100]$\% of all the obtained heterophilic nodes for neighbor manipulation.

The F1 and AUC-ROC of the four datasets are presented in \figurename~\ref{fig:FigTranslatedF1} and \ref{fig:FigTranslatedAUC}, respectively. For all the $p$ values, the performance of CAGAD consistently improves with the increase of translated neighbors on all datasets in terms of F1 and AUC-ROC.
The performance improvement is more significant for T-Finance and Amazon datasets when we increase the ratio of translated neighbors, as the ratios of anomalies to all neighbors in T-Finance and Amazon datasets being lower.
For example, the ratio on Amazon is only 6.8\%, while the ratio of anomalies on PubMed is around 20.8\%.
When there are only a few anomalies, for an anomalous node, the proportion between normal and anomalous neighbors is relatively high. Hence we need more anomalous neighboring nodes to balance the neighborhoods.
This observation demonstrates that our proposed counterfactual data augmentation is more effective on graphs with scarce anomalies.
Besides, when the selected neighbor ratio exceeds 70\% the performance keeps stable, which indicates that just a given number of anomalous neighboring nodes are required to obtain expected performance. 
Also, selecting more heterophilic nodes for neighbor manipulation ($p$) improves anomaly detection performance, which indicates that the heterophilic node detector can accurately identify nodes with heterophily-dominant neighbors, and manipulating them can benefit anomaly detection.

\subsection{The Effect of Detection Models}
\label{sec:DetectionModels}

\begin{figure}[t]
        \centering
        \subfloat[Detection performance using GCN model]{\label{fig:modelGCN}\includegraphics[width=0.47\textwidth]{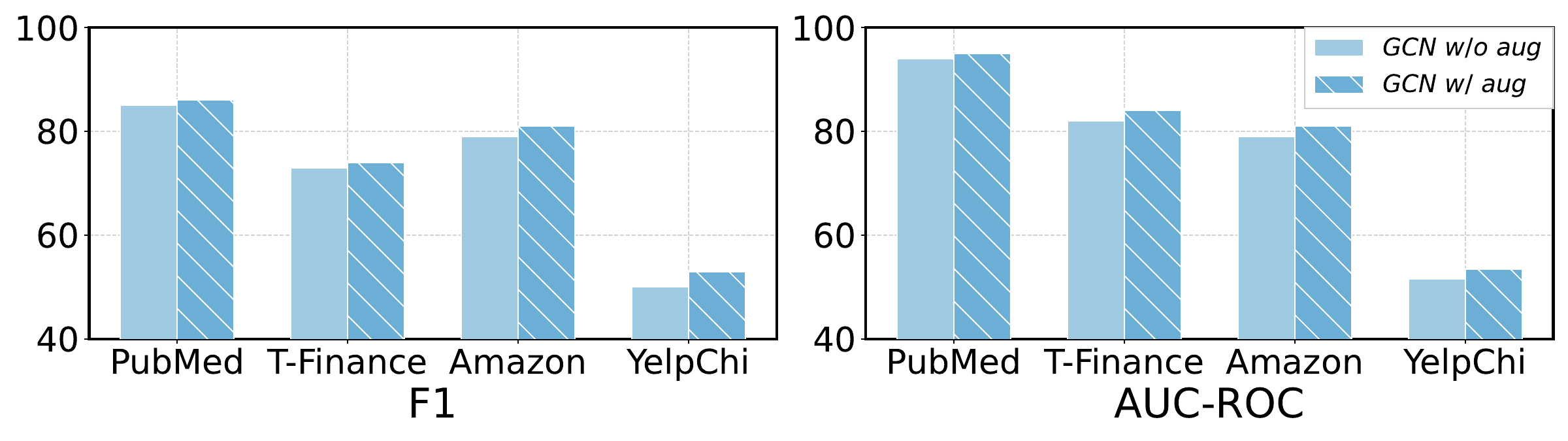}}

        \subfloat[Detection performance using GraphSAGE model]{\label{fig:modelGraphSAGE}\includegraphics[width=0.47\textwidth]{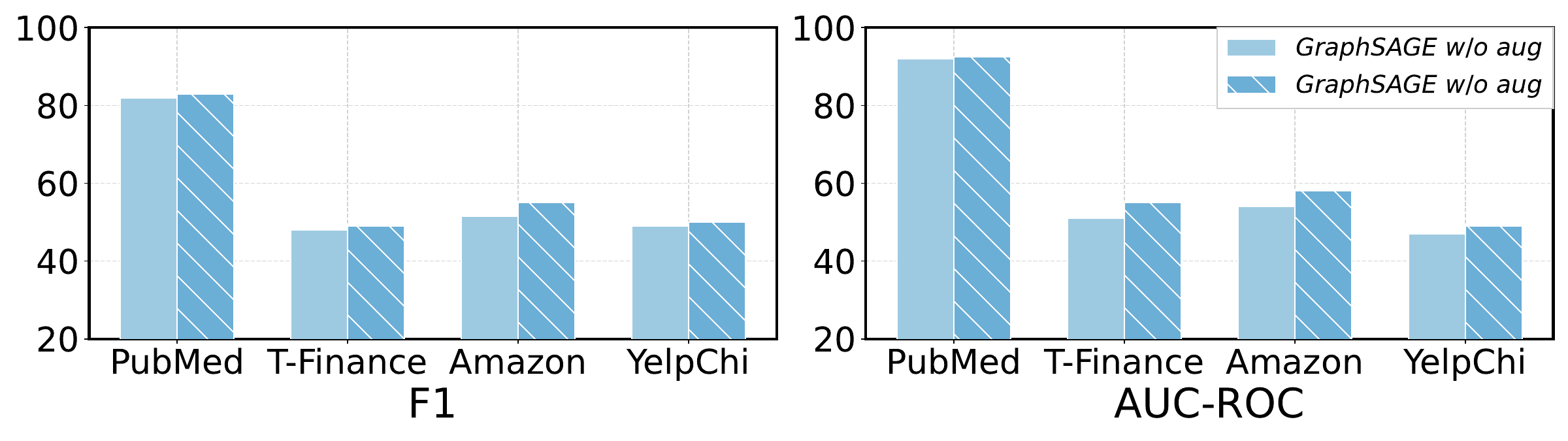}}

        \caption{The effect of the augmented data.}
        \label{fig:roleModel}
        \vspace{-3mm}
\end{figure}

We further investigate the effectiveness of the proposed counterfactual augmentation when applying it to other GNN-based detection methods: GCN and GraphSAGE.
We evaluate the model performance with and without the augmented data for graph anomaly detection and denote the two schemes as \textit{ModelName w/ aug} and \textit{ModelName w/o aug}. \figurename~\ref{fig:roleModel} shows the F1 and AUC-ROC for the GCN and GraphSAGE models on four datasets. The augmented data are generated by our designed DDPM-based generator. We 
observe that for both the GCN and GraphSAGE model, 
the one with the augmented data performs better. 
For example, the AUC of \textit{GCN w/ aug} is around 2.6\% higher than that of \textit{GCN w/o aug} on average. The 
results suggest that the augmented data generated by our method can improve the detection performance and 
the proposed counterfactual data augmentation strategy can be applied to other graph anomaly detection frameworks.

\subsection{Hyper-Parameter Sensitivity}
\label{sec:hyperPar}

\begin{figure}[t] 
\centering
\includegraphics[width=\linewidth]{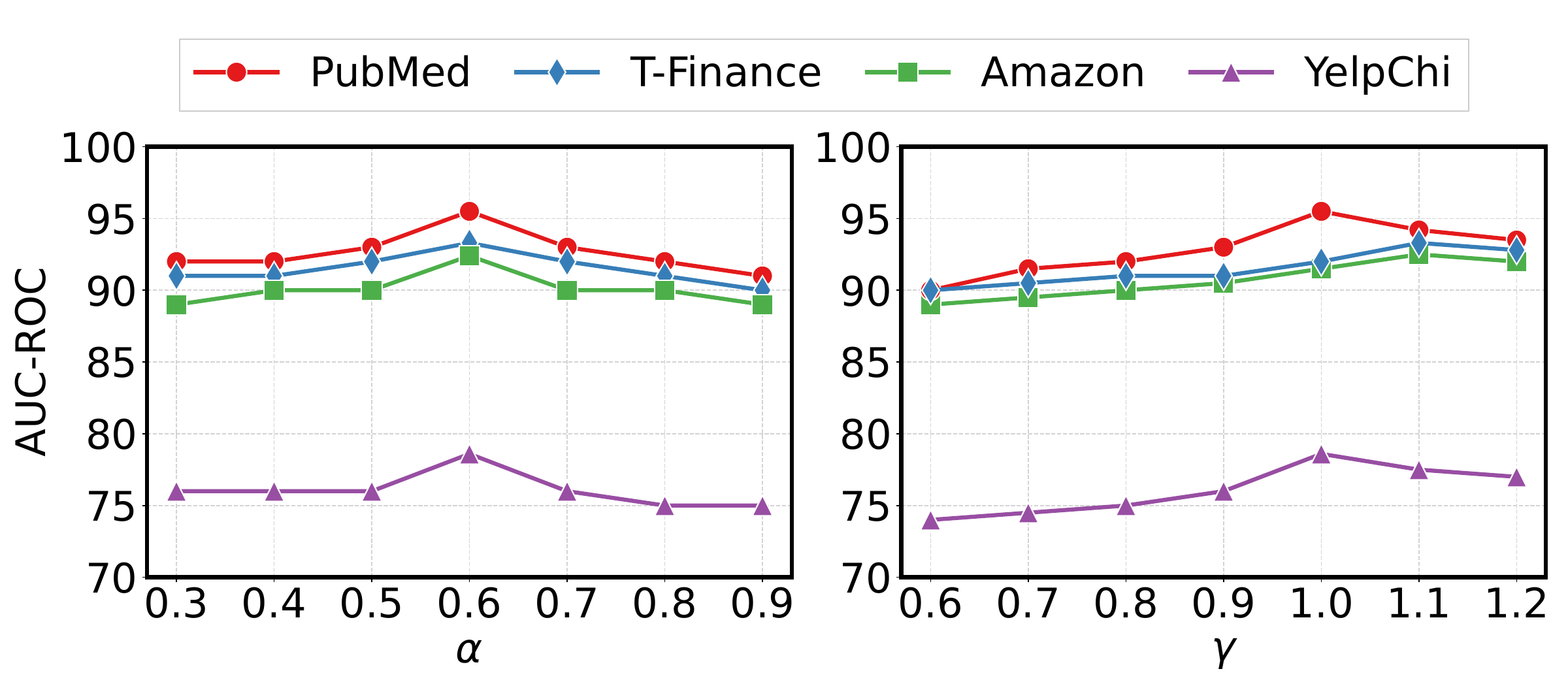}
\caption{The effect of the hyper-parameters.}
\label{fig:parameter}
\vspace{-3mm}
\end{figure}

We now evaluate the impact of two important hyper-parameters $\alpha$ and $\gamma$. In the heterophilic node detector, a 
heterophilic node 
is identified by comparing its heterophily degree $h_i^d$ with $\alpha$, which directly influences the results of the identification. As shown in the left of \figurename~\ref{fig:parameter}, the detection performance improves with the increase of $\alpha$ and then starts to decrease when $\alpha > 0.6$. When $\alpha$ is too small, many nodes might be incorrectly identified as heterophilic. 
In contrast, 
larger $\alpha$ prevents our designed heterophilic node detector from taking effect since too few nodes are identified. 
This indicates that a proper $\alpha$ can obtain expected identification results. The reason may be attributed to the fact that for anomaly detection tasks, there are only two categories, making the identification of heterophilic nodes feasible.
The weight parameter $\gamma$ in Eq.~\eqref{equ:OppFinal} 
is used to scale the different anomaly features for embedding generation. 
As shown in the right of \figurename~\ref{fig:parameter}, 
a higher value of $\gamma$ around $1.1$ is preferable to obtain a good performance since higher $\gamma$ enables the generator to excessively amplify the anomalous characteristics
in the generated embeddings.

\section{Related Work}
\label{sec:relate}


This work is mainly related to three research areas: graph anomaly detection, graph data augmentation, and diffusion probabilistic models. Here, we will present an overview of the most closely related works in each area, and highlight the major differences between our study and these works.

\subsection{Anomaly Detection on Graphs}
Since graph-structured data is 
ubiquitous and has 
capacity to model a wide range of real-world complex systems, 
identifying anomalies in graphs has drawn increased research interest \cite{gao2023addressing,he2022graph,li2024controlled}. 
Due to their 
demonstrated 
superior modeling power for graphs, various GNN-based methods 
have been proposed to detect anomalies on graphs.
The pioneer used GNNs to build an autoencoder to reconstruct the attribute and structure information simultaneously, and the abnormality is evaluated by reconstruction errors~\cite{ding2019deep}.
Based on this framework, a tailored deep GCNN is designed to detect local, global, and structural anomalies by capturing community structure in the graph~\cite{luo2022comga}.
Contrastive learning and self-supervised learning with GNNs are also introduced to identify the anomalies in attributed networks~\cite{zhang2022reconstruction,zheng2021generative}.
Meta-learning and hypersphere learning are incorporated into GNNs to leverage the labeled samples for anomaly detection \cite{ding2021few,2020DeepSAD,kumagai2021semi,zhou2022unseen}.

To remedy the problem that numerous neighbors with normal labels might make the anomaly representations learned by GNNs less distinguishable,
multiple re-sampling (e.g., over-sampling and under-sampling) strategies are designed in~\cite{dou2020enhancing,liu2021pick,liu2020alleviating}.
Researchers also utilized re-weighting methods to assign different weights to different samples \cite{wang2019semi,cui2020deterrent,liu2021intention}.
More recently, spectral filters are explored to enhance the expressive power of GNNs for learning better anomaly representations~\cite{tang2022rethinking,chai2022can}.

The re-sampling and re-weighting strategies 
generally require a number of labeled samples, and can only manipulate the training data for building a robust model and ignore the manipulation of test data where there still exists the phenomenon of imbalanced neighbors. We also aim to change the neighbors of anomalies to enhance node representation but
with a more proper neighborhood generation process for learning anomaly representations instead of sampling existing ones.

\subsection{Graph Data Augmentation}
These techniques aim to generate extra data by applying label-preserving transformations on inputs for improving the model's generalization capability.
A widely used scheme is related to edge operations, such as edge dropping and subgraph sampling~\cite{ding2022data,zhao2021data}.
One of the early works \cite{rong2020dropedge} randomly drops a fixed fraction of edges to generate new graph views for node classification.
Following this, task-irrelevant edges are identified and removed by the MLP-based graph sparsification model~\cite{zheng2020robust} and the nuclear norm regularization loss~\cite{luo2021learning} for improving the generalization performance.
These augmentation methods are also applied to contrastive learning and self-supervised learning for node classification and anomaly detection~\cite{zhu2021graph,liu2021anomaly}.
Recently, feature augmentation has been adopted for graph data augmentation, which aims to improve the node feature quality by learning additional task-relevant features.
Feature augmentation is generally utilized to initiate node features on plain graphs to smoothly incorporate into GNN models and supplement additional node features that are hard to capture by downstream models~\cite{liu2022local,kong2022robust}.
Besides, counterfactual data augmentation is exploited to improve the performance of node classification and link prediction~\cite{ma2022learning,zhao2022learning,xiao2023counterfactual}.
The researchers either generate counterfactual data by injecting interventions on the sensitive attribute of nodes~\cite{ma2022learning,xiao2023counterfactual} or find out counterfactual links which are the most similar node pairs with different treatments (neighborhood relations)~\cite{zhao2022learning} to boost the performance.

The above-mentioned graph data augmentation methods 
focus on increasing training set by creating label-preserving data or seeking existing counterfactual samples.
By contrast, we aim to alleviate the problem that GNNs produce indistinguishable representations for extremely imbalanced node distribution in anomaly detection. 
Correspondingly, our model translates normal nodes into the ones with opposite labels and steers GNN neighborhood aggregation to manufacture effective counterfactual data.
Also, our method can produce counterfactual data in an unsupervised way, which can be applied to test data
for enhancing node representations.

\subsection{Diffusion Probabilistic Models}
Diffusion probabilistic models aim to generate high-quality data samples by reversing the diffusion process via a Markov chain with discrete timesteps~\cite{ho2020denoising}, and 
have acquired state-of-the-art generation instances for many real-world applications. For example,
image generation is one of the major applications where diffusion models are exploited to produce high-quality images~\cite{song2021score,meng2022sdedit}, speed up the sampling process~\cite{xiao2022tackling,dockhorn2022score}, conduct image super-resolution~\cite{saharia2022image,li2022srdiff} and image-to-image translation~\cite{sasaki2021unit,sinha2021d2c}.
Waveform synthesis is another main application where the diffusion models are utilized to manufacture time-domain speech audio from the prior noise \cite{chen2021wavegrad,lam2022bddm}.
Diffusion models have also been applied to voice conversion~\cite{popov2022diffusion}, shape generation~\cite{zhou20213d}, and time series forecasting ~\cite{rasul2021autoregressive}.
Different from these methods that use diffusion models for regular data types (e.g., image and waveforms), we present a graph-specific diffusion model, which can iteratively extract distinct features of anomalies and exert them on the generative process to manufacture anomalous nodes.

\section{Conclusions}
\label{sec:conclusion}


In this paper, we proposed a counterfactual data augmentation method for graph anomaly detection, CAGAD, which can produce counterfactual data by steering the process of GNN neighborhood aggregation. In this framework, we designed a graph-specific diffusion model, which can translate a node embedding to the one with the anomalous label. This model is incorporated into GNN neighborhood aggregation to build a counterfactual GNN, which can learn distinguishable node representations to boost anomaly detection performance. The experiments on four real-world datasets show that CAGAD achieves state-of-the-art performance compared to many strong baselines, and the counterfactual graph augmentation strategy can be applied to other anomaly detection frameworks.

There are two limitations about our proposed model, which can become fruitful directions of further investigation. First, our counterfactual data augmentation method is only applicable to binary classification scenarios. Hence, we are interested in extending the model to accommodate situations involving multi-class classification. Second, we only investigate the effectiveness on static graphs. Thus, more evaluation is needed to understand the feasibility in the context of more complex graph data, e.g., heterogeneous graphs, spatial-temporal graphs, and dynamic graphs.

\section*{Acknowledgments}
This work is supported by the National Natural Science Foundation of China (Grant No. 62176043 and 62072077).

\bibliographystyle{IEEEtran}
\bibliography{sample-base}

\begin{IEEEbiography}[{\includegraphics[width=1in,height=1.25in,clip,keepaspectratio]{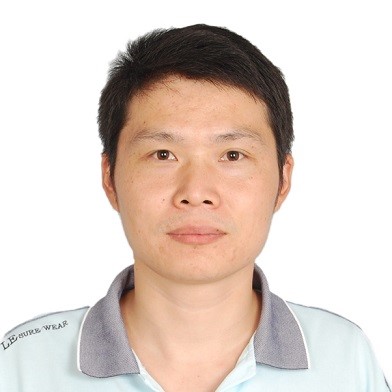}}]{Chunjing Xiao} received the Ph.D. degree in computer software and theory from the University of Electronic Science and Technology of China, Chengdu, China, in 2013.

He is currently an Associate Professor with the
School of Computer and Information Engineering,
Henan University, Kaifeng, China. He was a
Visiting Scholar with the Department of Electrical
Engineering and Computer Science, Northwestern
University, Evanston, IL, USA. His current research
interests include  anomaly detection, recommender systems, and Internet of Things.
\end{IEEEbiography}

\begin{IEEEbiography}[{\includegraphics[width=1in,height=1.25in,clip,keepaspectratio]{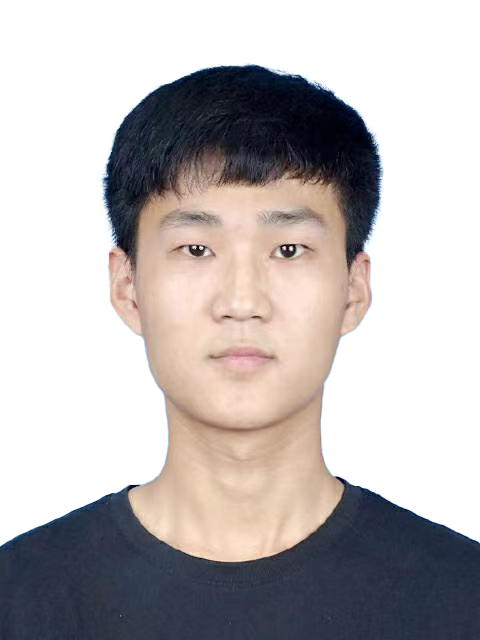}}]{Shikang Pang} received the B.E degree from
the School of Computer and Information Engineering, Huanghuai University, China, in 2019.
He is currently pursuing the M.S. degree in the
School of Computer and Information Engineering, Henan University.

His current research interests include anomaly detection, data analytics and social network data mining.
\end{IEEEbiography}

\begin{IEEEbiography}[{\includegraphics[width=1in,height=1.25in,clip,keepaspectratio]{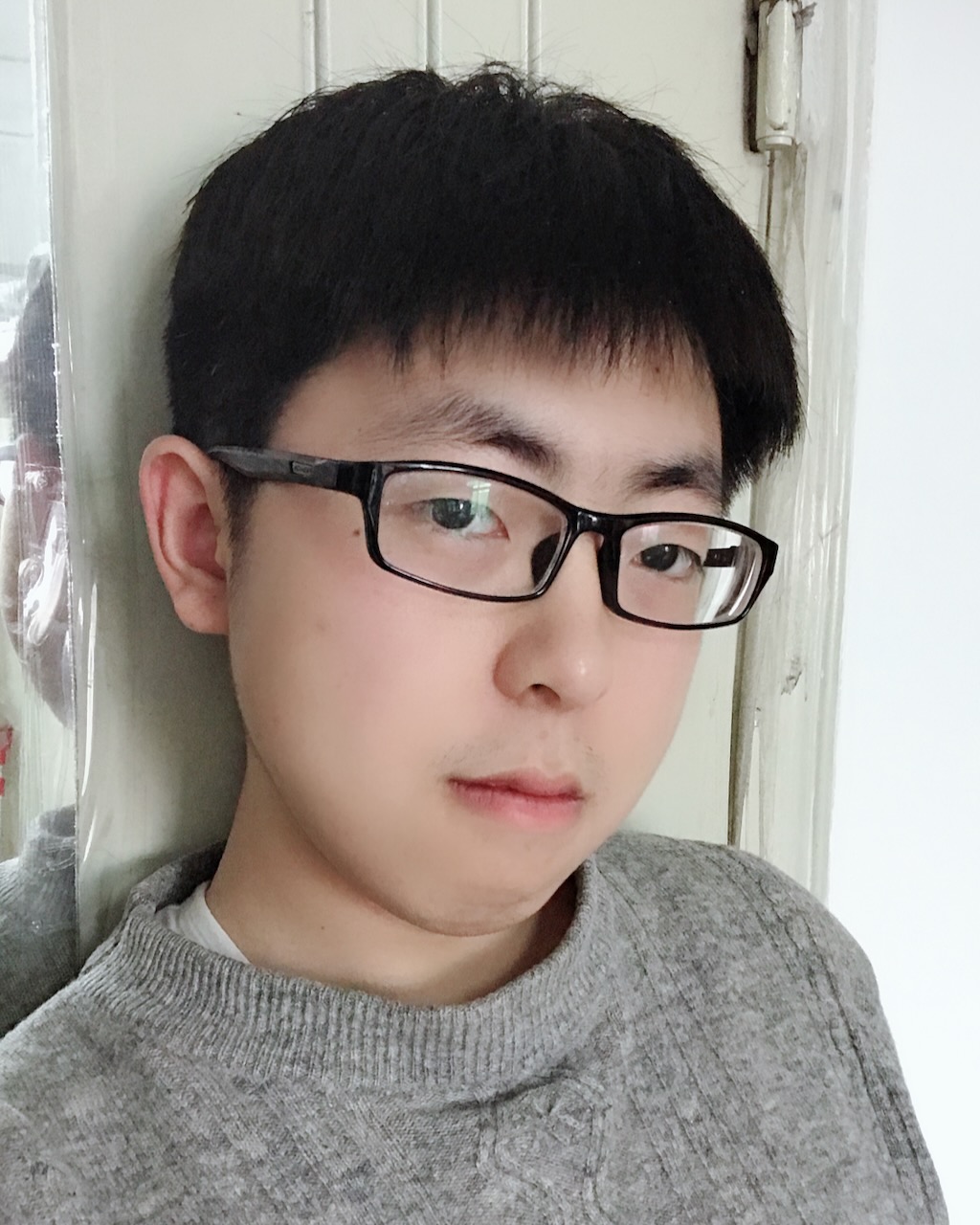}}]{Xovee Xu} received the B.S. degree and M.S. degree in software engineering from the University of Electronic Science and Technology of China (UESTC), Chengdu, Sichuan, China, in 2018 and 2021, respectively. He is currently pursuing the Ph.D. degree in computer science at UESTC.

His recent research interests include social
network data mining and knowledge discovery,
primarily focuses on information diffusion in full-scale graphs, human-centered data mining, representation learning, and their novel applications in various social and scientifc scenarios.
\end{IEEEbiography}

\begin{IEEEbiography}[{\includegraphics[width=1in,height=1.25in,clip,keepaspectratio]{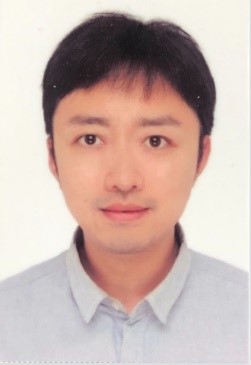}}]{Xuan Li} received the B.S degree and M.S degree in software engineering from the Sichuan University, Chengdu, Sichuan, China in 2009 and 2012, respectively. He is currently pursuing the Ph.D. degree in electronic information at Sichuan University. Meanwhile he works as a teacher at Sichuan Post and Telecommunication College.

His current research interests include anomaly detection, data analytics, social network data mining and knowledge discovery.
\end{IEEEbiography}

\begin{IEEEbiography}[{\includegraphics[width=1in,height=1.25in,clip,keepaspectratio]{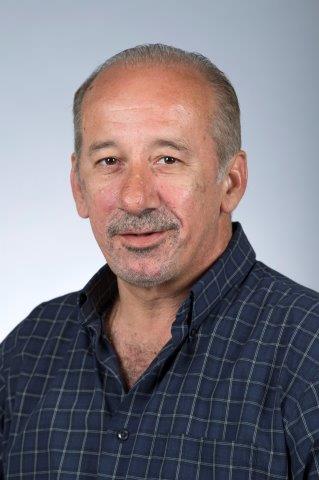}}]{Goce Trajcevski} (Member, IEEE) received the B.Sc. degree in informatics and automation from the University of Sts. Kiril i Metodij, Skopje, North Macedonia, in 1989, and the M.S. and Ph.D. degrees in computer science from the University of Illinois at Chicago, Chicago, IL, USA, in 1995 and 2002, respectively.

He is currently an Associate Professor with the Department of Electrical and Computer Engineering, Iowa State University, Ames, IA, USA. His research has been funded by the NSF, ONR, BEA, and Northrop Grumman Corporation. In addition to a book chapter and three encyclopedia chapters, he has coauthored over 140 publications in refereed conferences and journals. His main research interests are in the areas of spatiotemporal data management, uncertainty and reactive behavior management in different application settings, and incorporating multiple contexts.

Dr. Trajcevski was the General Co-Chair of the IEEE International Conference on Data Engineeing 2014 and ACM SIGSPATIAL 2019, the PC Co-Chair of the ADBIS 2018 and ACM SIGSPATIAL 2016 and 2017, and has served in various roles in organizing committees in numerous conferences and workshops. He is an Associate Editor of the ACM Transactions on Spatial Algorithms and Systems and the Geoinformatica Journals.
\vspace{-1cm}
\end{IEEEbiography}

\begin{IEEEbiography}[{\includegraphics[width=1in,height=1.25in,clip,keepaspectratio]{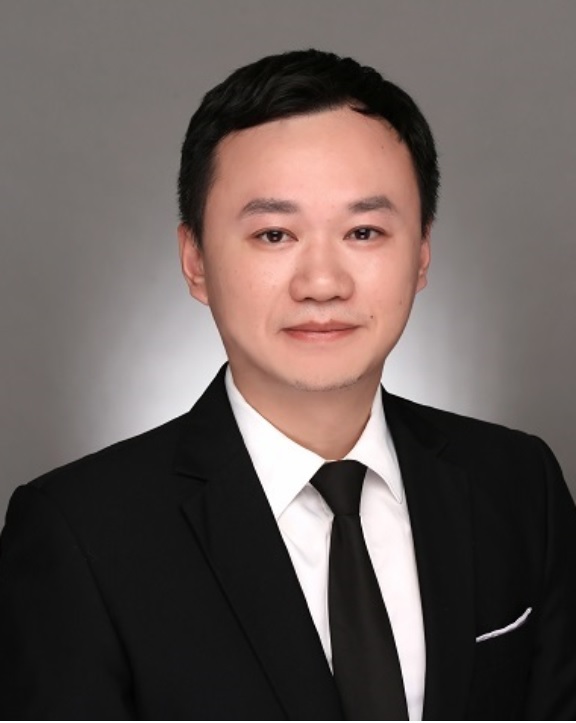}}]{Fan Zhou} (Member, IEEE) received the B.S. degree in computer science from Sichuan University, Chengdu, China, in 2003, and the M.S. and Ph.D. degrees in computer science from the University of Electronic Science and Technology of China, Chengdu, in 2006 and 2012, respectively.

He is currently a Professor with the School of
Information and Software Engineering, University
of Electronic Science and Technology of China. His
research interests include machine learning, neural
networks, spatio-temporal data management, graph
learning, recommender systems, and social network data mining.
\vspace{-1cm}
\end{IEEEbiography}

\end{document}